# Linguistic Changes in Spontaneous Speech for Detecting Parkinson's Disease Using Large Language Models

**Jonathan Crawford[1,2]**

[1]University School of Milwaukee, Milwaukee, Wisconsin, USA
[2]Department of Electrical and Computer Engineering, Boston University, Boston, Massachusetts, USA

**\*Correspondence:**
Jonathan Crawford
jcraw@bu.edu

## Abstract

Parkinson's disease is the second most prevalent neurodegenerative disorder with over ten million active cases worldwide and one million new diagnoses per year. Detecting and subsequently diagnosing the disease is challenging because of symptom heterogeneity with respect to complexity, as well as the type and timing of phenotypic manifestations. Typically, language impairment can present in the prodromal phase and precede motor symptoms suggesting that a linguistic-based approach could serve as a diagnostic method for incipient Parkinson's disease. Additionally, improved linguistic models may enhance other approaches through ensemble techniques. The field of large language models is advancing rapidly, presenting the opportunity to explore the use of these new models for detecting Parkinson's disease and to improve on current linguistic approaches with high-dimensional representations of linguistics. We evaluate the application of state-of-the-art large language models to detect Parkinson's disease automatically from spontaneous speech with up to 73% accuracy.

## 1	Introduction

Parkinson's disease (PD) is the second most prevalent neurodegenerative disorder with over ten million active cases worldwide, one million new diagnoses per year, and an exponentially growing incidence rate (Ou et al., 2021). The global prevalence of PD continues to rise due to increased life expectancy and industrialization (Dorsey et al., 2018). PD is a chronic and progressive neurodegenerative disorder that induces physical and cognitive impairment. The pathogenesis and pathophysiology of the disease is poorly understood. Current research indicates that the pathogenesis of PD involves an interplay of unknown genetic susceptibilities and environmental exposures (Kouli et al., 2018). PD is characterized by the progressive loss of dopaminergic neurons in the substantia nigra and the presence of Lewy bodies, leading to central nervous system degradations. The disease is also characterized by motor symptoms, namely bradykinesia, resting tremor, rigidity, and postural instability. Non-motor symptoms also manifest heterogeneously (Simuni and Sethi, 2009).

PD is diagnosed late relative to the pathogenesis and with low accuracy (Rizzo et al., 2016; Beach and Adler, 2018). This can be attributed to multiple confounding factors including the absence of early-stage biomarkers and screening methods, the complex symptomatology of the disease, and the limitations of diagnostic methods in timely detection and differentiation. The pathogenesis of the disease is estimated to begin decades before the manifestation of the phenotypic symptoms necessary for clinical diagnosis (Kilzheimer et al., 2019). PD is currently diagnosed with clinical evaluations. The clinical evaluation utilizes phenotypic symptoms augmented with neuroimaging to exclude other conditions. The primary symptoms used for PD diagnosis are tremor, rigidity, bradykinesia, and postural instability. These fine motor-skill deteriorations are considered the cardinal and first observable signs of PD. Dependence on these symptoms is problematic given their variability, non-specificity, and potential overlap with other diseases (Adler et al., 2014). Inconsistent symptom onset and presentations across populations further adds to the complex symptomatology. Reliance on these physical symptoms leads to late detection because these symptoms do not present until around 80% neural degradation (Bernheimer et al., 1973).

There is a need for additional biomarkers and new methods to detect PD. Language impairment can present in the prodromal phase and precede motor symptoms suggesting that a linguistic-based approach could serve as a diagnostic method for incipient PD (Postuma and Berg, 2019). Linguistics models may also be used to detect PD across all stages and enhance other approaches through ensemble techniques.

The architecture, parameter structure, and training of the large language models can be leveraged to extract and encode into text embeddings a unique feature space representing the morphology, syntax, semantics, and pragmatics of the spontaneous speech signals. For example, Bidirectional Encoder Representations from Transformers (BERT) has been used to detect PD (Devlin et al., 2018). The field of large language models is advancing rapidly, which presents the opportunity to explore the use of the new models for detecting PD and to improve on current linguistic approaches with high-dimensional representations of linguistics.

We evaluate the application of state-of-the-art large language models to detect PD from spontaneous speech. The state-of-the-art large language models lead to improved performance over the prior methods using our implementation.

## 2    Material and Methods

A high-level description of a system to detect PD is detailed in Figure 1. The system inputs digitized spontaneous speech from participants. The participants either belong to a control group without PD or have PD of varying degrees of severity. The speech is transcribed automatically with an automated speech recognition model. A large language model then generates a linguistic feature space from the transcription. A classification algorithm subsequently processes the feature space to make a PD /Non-PD diagnosis. Low-level details on each step are presented below. Implementation details can be found in the appendix; the computer code used to create the results shown in this paper is available upon request from the author.

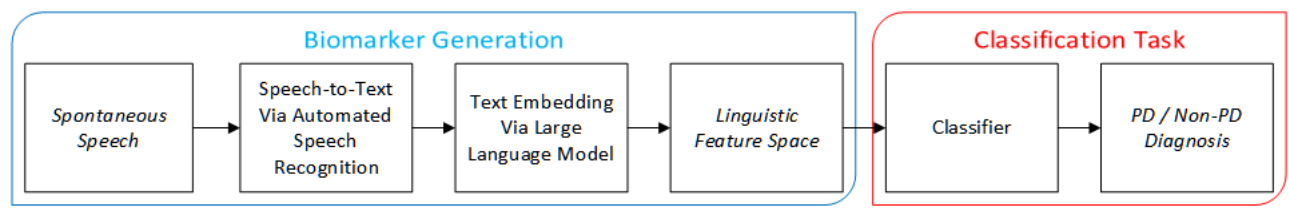

**Figure 1.** A high-level representation of the methodology. Italicized boxes signify inputs and outputs.

## 2.1 Dataset (Spontaneous Speech)

The PC-GITA dataset is used in this study (Orozco-Arroyave et al., 2014). The dataset consists of the spontaneous speech of 50 subjects with PD and 50 health controls (HC). The data is age and gender matched. The p-value for age, calculated with a two-sided Mann-Whitney U test, is 0.99. The data consist of monologues where each subject is asked to speak about what they do on a normal day. The speech is of native Colombian Spanish speakers. The recordings were taken with the PD patients in the ON state, which refers to a period of time when medication effectively alleviates the motor symptoms of the disease. Recordings were conducted no more than three hours after medication was taken. The healthy controls do not have symptoms associated with PD or any other neurological disease. Data for each PD participant is labeled with Movement Disorder Society-Unified Parkinson's Disease Rating Scale Part III (MDS-UPDRS-III), Hoehn & Yahr (H&Y), and time after diagnosis. Data for both the HC and PD groups are labeled with sex and age.

**Table 1.** Patient demographics and calculated statistics derived from the dataset. μ: average, σ: standard deviation.

|  | PD Patients | | HC Subjects | |
| --- | --- | --- | --- | --- |
|  | Male | Female | Male | Female |
| Number of subjects | 25 | 25 | 25 | 25 |
| Age [years] (μ±σ) | 61.3 ± 11.4 | 60.7 ± 7.3 | 60.5 ± 11.6 | 61.4 ± 7.0 |
| Range of age [years] | 33 - 81 | 49 - 75 | 31 - 86 | 49 - 76 |
| Time post diagnosis [years] (μ±σ) | 8.7 ± 5.8 | 13.8 ± 12.4 |  |  |
| Range of time post diagnosis [years] | 0.4 - 20 | 1 - 43 |  |  |
| MDS-UPDRS-III (μ±σ) | 37.8 ± 22.1 | 37.6 ± 14.0 |  |  |
| Range of MDS-UPDRS-III | 6 - 93 | 19 - 71 |  |  |

## 2.2 Biomarker Generation

### 2.2.1 Speech-to-Text Via Automated Speech Recognition

The audio files are automatically transcribed into textual form using Whisper, an automated, multilingual speech recognition model from Open AI (Radford et al., 2022). The transcription endpoint is used to transcribe the audio into Spanish.

### 2.2.2 Text Embedding Via Large Language Model

Text embeddings are generated from the transcriptions using large language models. We evaluate the efficacy of multiple state-of-the-art large language models. Each model generates a high dimensional linguistic feature space with each dimension representing different linguistic features determined by the architecture and training of the models. The models were chosen because they are widely applied and considered to be state-of-the-art as of the writing of this paper.

We present results from: Bidirectional Encoder Representations from Transformers (BERT) (Devlin et al., 2018); XLNet (Yang et al., 2019); Generative Pre-trained Transformer 2 (GPT-2) (Radford et al., 2019); text-embedding-ada-002 (Greene et al., 2022; Neelakantan et al., 2022); and text-embedding-3-small and text-embedding-3-large (OpenAI, 2024). We note that BERT has been previously applied for PD detection (Devlin et al., 2018).

The dimensionality of the text-embedding-3-small and text-embedding-3-large outputs can be reduced through an API endpoint parameter. This represents a trade-off between performance and the cost of using embeddings. Specifically, embeddings are shortened internally by the model without losing their concept-representing properties (Kusupati et al., 2022). We present results from the two models at both their default (maximum) dimensionality and at a reduced dimensionality, adjusted to match that of the other evaluated models. This approach enables a direct comparison of performance, standardized based on dimensionality, between these models and the other models evaluated.

## 2.3 Classification

The classification task presents challenges due to the high dimensionality of the feature spaces compared to the small size of the dataset. A support vector machine (SVM) is utilized for its robustness when handling such high-dimensional data and for its effectiveness in classification tasks through the creation of optimal hyperplanes in a transformed feature space. Specifically, SVM models are resistant to overfitting when regularized (Xu et al., 2008). All dimensions of the embeddings are used to train the SVM.

Assessing performance is challenging due to the limited dataset sizes. To address this, 10-fold stratified cross-validation is employed to tune hyperparameters and provide a stochastic estimate of performance. The dataset consists of 100 samples, so 90 are used as the training set and 10 are treated as an unseen and validation set in each fold. A grid search is used to optimize the hyperparameters based on accuracy. The hyperparameters considered are the kernel, the regularization parameter (C), and the kernel coefficient gamma. The kernel types considered are Polynomial, Radial Basis, and Sigmoid functions, which are denoted in the results section below as poly, rbf, and sig, respectively. The regularization parameter ranges over the values $[10^{-5}, 10^{-4}, \ldots, 10^{5}]$, which helps in controlling the trade-off between achieving a low training error and a low testing error, thereby avoiding overfitting. The kernel coefficient, which influences the decision boundary, also spans across the same range of values: $[10^{-5}, 10^{-4}, \ldots, 10^{5}]$.

After each fold in the cross-validation process, the evaluation metrics are recorded, and their mean values are calculated upon completion of all folds. The evaluation metrics are accuracy, precision, recall, and the area under the curve (AUC). Positives are defined as the group with PD and negatives are defined as the healthy control group. Accuracy is defined as correct predictions over all predictions. Precision is defined as true positives over positive predictions. Recall is defined as true positives over all positives. Area under the curve is defined as area under the receiver operating characteristic curve. Standard Error is defined as the standard deviation of the metric across folds divided by the square root of the number of samples, ten in this case.

## 3   Results

**Table 2.** Results of the application of state-of-the-art large language models. Performance metrics and standard errors are reported as percentages, expressed as mean (μ) ± standard error (SE).

| Embedding Model | Dimension | Accuracy | Precision | Recall | AUC | Kernel | C | Gamma |
|---|---|---|---|---|---|---|---|---|
| BERT | 768 | 66 ± 4.8 | 78 ± 10.7 | 44 ± 7.2 | 59 ± 5.3 | rbf | $10^{-1}$ | $10^{-1}$ |
| XLNet | 768 | 64 ± 3.1 | 64 ± 4.7 | 64 ± 5.8 | 66 ± 3.8 | rbf | $10^{1}$ | $10^{-5}$ |
| GPT-2 | 768 | 67 ± 3.7 | 66 ± 4.8 | 66 ± 7.3 | 70 ± 6.8 | poly | $10^{-5}$ | $10^{-3}$ |
| text-embedding-ada-002 | 1536 | 70 ± 3.0 | 80 ± 5.8 | 60 ± 5.2 | 76 ± 5.1 | sig | $10^{1}$ | $10^{-1}$ |
| text-embedding-3-small | 1536 | 73 ± 2.6 | 80 ± 4.9 | 68 ± 6.8 | 78 ± 2.7 | poly | $10^{-4}$ | $10^{1}$ |
| text-embedding-3-small | 768 | 72 ± 2.5 | 77 ± 4.4 | 68 ± 6.8 | 78 ± 2.8 | poly | $10^{-4}$ | $10^{1}$ |
| text-embedding-3-large | 3072 | 73 ± 3.7 | 71 ± 4.5 | 80 ± 6.0 | 76 ± 4.6 | rbf | $10^{-4}$ | $10^{-5}$ |
| text-embedding-3-large | 768 | 71 ± 4.3 | 74 ± 5.6 | 68 ± 6.1 | 74 ± 5.7 | rbf | $10^{2}$ | $10^{-2}$ |

## 4 Discussion

We have demonstrated that the state-of-the-art large language models can detect PD with up to 73% accuracy using a linguistic feature space generated with large language models. We show that the text-embedding-3 models, outperform the other models. This finding is consistent with the benchmarked performance of all of the models across a variety of tasks (Muennighoff et al., 2022). The previous research for PD detection with large language models, specifically BERT, is only 66% accurate with our implementation and with the dataset that we used (Escobar-Grisales et al., 2023). We demonstrate that the text-embedding-3 models surpass BERT across performance benchmarks.

The performance metrics for text-embedding-3 are largely independent of the dimensionality of the embedding output. In particular, even with the dimensionality reduced to 768 to match BERT, the performance metrics are still better. Therefore, we conclude that the better performance with the use of the state-of-the-art models is due to the intrinsic architecture of the large language models and not due to the increased dimensionality.

Comparison between different detection methods is difficult due to the use of proprietary datasets and insufficient implementation details. We aim to overcome these limitations by using a dataset that is in the public domain, providing implementation details, and making our source code available upon request. This enables other researchers to replicate our methods and implementation.

### 4.1 Past Work

Motor-speech impairment is estimated to occur in over 90% of PD cases (Ramig et al., 2011). Past research has shown that acoustic features extracted from speech signals including prosodic, vocal, and lexical elements can be used to detect PD (Dixit et al., 2023). Research has also been conducted on the processing of acoustic speech signals for detecting Alzheimer's disease (Luz et al., 2021).

Acoustic models of speech have demonstrated high accuracy on performance tasks. However, acoustic features are phonetic and arise from physical changes in the vocal tract. They therefore may not manifest until late into the pathogenesis, which limits the utility of acoustic models for screening. Moreover, acoustic models rely on myriad acoustic cues, as no single cue provides a robust enough feature space (Basak et al., 2023). Specific acoustic features are susceptible to noise and environmental sensitivities. Additionally, acoustic models may be more prone to overfitting on training data due to the curse of dimensionality. This can be attributed to their reliance on 20-30 acoustic cues for small datasets. The combination of these factors may reduce the robustness and specificity of the models and limit their potential for clinical implementation.

Linguistic models have also been proposed for both PD and Alzheimer's disease. Large language models have been utilized to distinguish Alzheimer's disease from spontaneous speech with 74% accuracy (Agbavor and Liang, 2022). However, the research is conducted in the context of an aphasia exam, and Alzheimer's disease and PD have different manifestations of language impairment. BERT has been implemented to detect PD from spontaneous speech (Escobar-Grisales et al., 2023). A framework for automated semantic analyses of action stories capturing action-concept markers was developed to distinguish PD (García et al., 2022). Morphological analysis tools were used to study cognitive impairment and utterance alterations in PD on a Japanese dataset (Yokoi et al., 2023). The research showed that cognitively unimpaired PD patients exhibited different usage rates of morphological language components when compared to the healthy control group. Morphological analysis tools encompass only one of the five linguistic components, whereas text embeddings capture four, including morphology, syntax, semantics, and pragmatics.

Research reports accuracy of up to 72% accuracy with the PC-GITA dataset using Word2Vec word embeddings on a manual transcription of the monologues (Pérez-Toro et al., 2019). Word embeddings models are context-independent, meaning each word has the same embedding regardless of its context in a sentence. Therefore, word embeddings primarily extract features at the word level. The study also eliminates punctuation and stop words and performs lexicon normalization. This process results in the loss of linguistic information beyond the semantic component of each isolated word. Manual transcription requires domain- and task-specific knowledge from the transcriber. The transcriber and their domain-experience therefore become a variable in the experiment and may introduce implicit bias during the transcription process. In contrast, automatic speech recognition models do not introduce the variable of the transcriber and provide a uniform and scalable approach. However, automatic speech recognition models may struggle with domain-specific jargon, accents, and speech nuances. Specifically, they may fail to identify, either by commission or omission, the explicit incorrect use of language that may be present in individuals with language impairments. Automated transcription may lower accuracy in performance tasks, specifically in the classification of neurodegenerative diseases (Soroski et al., 2022). We report the performance metrics of our text embeddings approach, applied to an automatically transcribed version of the monologues. We believe that the accuracy does not fully represent the capabilities of the approach. If the embedding models were applied to a manually transcribed version of the monologues, we may achieve even better performance.

## 4.2   Limitations

The dataset is small relative to the high dimensionality of the feature spaces, which may restrict the generalizability of the results. The small dataset size also increases the risk of overtraining the model and makes it challenging to create representative validation and testing sets. However, the SVM with regularization prevents overtraining. Additionally, the small number of samples in test sets limits the implementation of statistical methods to assess significance.

The potential for misdiagnosis in the PD patient group and undiagnosed neurological conditions in the control group introduces uncontrolled variables that could impact the findings of the study. The dataset also does not account for variations in dialects and accents, factors that could influence the performance of the model. For generating the monologues, participants were asked to describe their daily routine. The nature of the prompt may not induce the participants to exhibit all forms of language impairment present.

The time after diagnosis for PD patients ranges from 0.4 to 43 years. The high variability in disease duration may lower the application of the method for early disease detection. However, we note that the mean average time post diagnosis for the 50 PD patients is $11.2 \pm 9.9$ years. This indicates a skew towards the earlier phase of the disease. Additionally, 46% of the MDS-UPDRS-III scores within the PD cohort fall below the 32-point threshold. This indicates only mild motor impairment. 76% of the scores are beneath the 52-point threshold for severe motor impairment. This highlights that the majority of the cohort exhibits only mild to moderate symptoms.

The publication of "Attention is All You Need" (Vaswani et al., 2017) and the release of OpenAI's GPT-3 model has led to rapid growth in the fields of generative AI and large language models. Large language models may be implemented in a clinical setting to help detect PD in the future. This raises the concern of implementing a test based on a non-transparent and non-open-source model. Additionally, research has shown that large language models may reflect societal biases, including biases related to gender and race. This concern highlights a larger issue as large language models continue to be implemented in the healthcare field and society (Ghassemi et al., 2023). More research is needed to ensure that the models and tests are developed and implemented responsibly.

### 4.3 Future Directions

We recommend that future data collection efforts consider various forms of participant prompting. Examples of different prompts include memory-dependent tasks, narrative construction, abstract thinking, and problem-solving scenarios. Additionally, we suggest considering different mediums for conversational tasks, including monologue, dialogue, and multilogue formats. We recommend future data collection to incorporate a dimension of time. By tracking patients longitudinally, research could capture the progression of the disease and its linguistic markers. This could offer information on how early these markers appear and how they evolve. We also recommend the procurement of a standardized dataset with multiple neurodegenerative disorders and languages.

This research highlights that spontaneous speech as a classifiable biomarker through linguistic representation in text embeddings. Based on this observation, several questions and future directions emerge. Previous research has demonstrated high accuracy in using speech signals to distinguish various neurodegenerative diseases (Hecker et al., 2022). However, most of the research uses healthy controls and positives of the disease for binary classification. It remains unclear whether each neurodegenerative disease has a unique signature in its speech patterns, whether acoustic or linguistic. This raises the question of how these models will perform in the real world, where individuals may have other conditions with similar or overlapping symptoms. This also raises the question of whether a single model, either binary or multimodal, can distinguish between multiple neurodegenerative diseases or diseases with similar presentations. Text embeddings and other components of the speech signal could potentially be utilized to infer rating scale scores and other cognitive test results. There is therefore the possibility of developing a regression model capable of automatically assessing PD progression based on speech data. The manifestation of language impairments across various languages is still unclear. The Spanish-based model may be extended and applied to other languages. Techniques such as transfer learning and zero-shot learning may offer effective adaptation strategies for new languages. There also remains the possibility of increasing the accuracy and performance of our approach in this classification task by utilizing more complex methods such as computational Neural Networks and Deep learning. Long Short-Term Memory (LSTM) and Recurrent Neural Networks (RNN) are considered effective models for distilling and classifying text embedding feature spaces and may be applied to increase accuracy. An ensemble method incorporating acoustic and linguistic features may also be developed to increase performance.

## 5   Conflict of Interest

The author declares that the research was conducted in the absence of any commercial or financial relationships that could be construed as a potential conflict of interest.

## 6   Author Contributions

All work was conducted by the author Jonathan Crawford.

## 7   Acknowledgments

The author thanks the following people for supporting this project: Professor Juan Rafael Orozco-Arroyave for providing access to the PC-GITA corpus; Dr. Pippa Simpson, Dr. Henry Medeiros, Dr. Malcolm Slaney, and Dr. David Nahamoo for their technical expertise and guidance in developing the methodology; Dr. Greg Marks, Mr. Eliot Scheuer, and Mr. Bryan Pack for their continual support; and Dr. Michael Bartl for teaching the mathematics and problem-solving skills necessary to tackle this problem. Finally, the author is especially grateful for and thankful to Mr. Robert Juranitch for his persisting mentorship, and guidance.

## 8   Data Availability Statement

The dataset analyzed for this study is available on request from (Orozco-Arroyave et al., 2014). The computer code used to create the results shown in this paper is available upon request from the author.

## 9   Appendix: Implementation Details

The details of the implementation of the methods are shown in this section.

The code was written in Python using Google Colab.

The details of the Speech-to-Text Via Automated Speech Recognition step are as follows.

- Version: Whisper-1
- API Endpoint: https://api.openai.com/v1/audio/transcriptions
- Response_format parameter: set as "text"
- Other Parameters: none set or invoked

The details of the Text Embedding Via Large Language Model step are as follows.

OpenAI Endpoint Models:

- text-embedding-ada-002
    - API Endpoint: discontinued
    - Parameters:
        - model="text-embedding-ada-002"
- text-embedding-3-small
    - API Endpoint: https://api.openai.com/v1/embeddings
    - Parameters:
        - model="text-embedding-3-small"
        - dimensions=768 (for text-embedding-3-small with reduced dimensions; otherwise, the parameter is not invoked, and the dimension is automatically set to the default maximum length).
- text-embedding-3-large
    - API Endpoint: https://api.openai.com/v1/embeddings
    - Parameters:
        - model="text-embedding-3-large"
        - dimensions=768 (for text-embedding-3-large with reduced dimensions; otherwise, the parameter is not invoked, and the dimension is automatically set to the default maximum length).

Hugging face Transformers Endpoint Models: https://huggingface.co/docs/transformers/index

- BERT
    - Tokenizer: BertTokenizer.from_pretrained('bert-base-uncased')
    - Model: BertModel.from_pretrained('bert-base-uncased')
- XLNet
    - Tokenizer: XLNetTokenizer.from_pretrained('xlnet-base-cased')
    - Model: XLNetModel.from_pretrained('xlnet-base-cased')
- GPT-2
    - Tokenizer: GPT2Tokenizer.from_pretrained('gpt2')
    - Model: GPT2Model.from_pretrained('gpt2')

The performance metrics and SVM classifier are implemented with scikit-learn (Pedregosa et al., 2012).